\documentclass{article}

\usepackage{PRIMEarxiv}

\usepackage[utf8]{inputenc} 
\usepackage[T1]{fontenc}    
\usepackage{hyperref}       
\usepackage{url}            
\usepackage{booktabs}       
\usepackage{amsfonts}       
\usepackage{nicefrac}       
\usepackage{microtype}      
\usepackage{lipsum}
\usepackage{fancyhdr}       
\usepackage{graphicx}
\usepackage{multirow}
\usepackage{xcolor}

\usepackage{amsmath}
\usepackage{comment}

\graphicspath{{media/}}     

\title{Evaluation of Semantic Search and its Role in Retrieved-Augmented-Generation (RAG) for Arabic Language
}

\author{
  Ali Mahboub, Muhy Eddin Za'ter, Bashar Al-Rfooh, Yazan Estaitia, Adnan Jaljuli, Asma Hakouz  \\
  Maqsam \\
  Amman, Jordan\\
  \texttt{\{ali.mahbob, muhyeddin, bashar.alrfooh, yazan.estaitia, adnan.jaljuli, asma\}@maqsam.com} \\
}

\begin{document}

\maketitle

\begin{abstract}
The latest advancements in machine learning and deep learning have brought forth the concept of semantic similarity, which has proven immensely beneficial in multiple applications and has largely replaced keyword search. However, evaluating semantic similarity and conducting searches for a specific query across various documents continue to be a complicated task. This complexity is due to the multifaceted nature of the task, the lack of standard benchmarks, whereas these challenges are further amplified for Arabic language. This paper endeavors to establish a straightforward yet potent benchmark for semantic search in Arabic. Moreover, to precisely evaluate the effectiveness of these metrics and the dataset, we conduct our assessment of semantic search within the framework of retrieval augmented generation (RAG).
\end{abstract}


\section{Introduction}

The abundance of information has driven the development of semantic search technologies that surpass traditional keyword-based search engines by understanding the context and intent of user queries through natural language processing (NLP) and machine learning \cite{huang2013learning}. Unlike conventional search methods that focus on matching keywords, semantic search interprets the meaning and relationships between words, aiming to mimic human understanding. This advancement enhances user experience across various applications, including web search engines, knowledge discovery, and personalized content recommendation systems, and most recently Retriever-Augmented Generation (RAG) \cite{gavankar2020comparative}.

RAG represents an innovative approach at the crossroads of information retrieval and natural language generation, leveraging the strengths of both fields to refine Artificial Intelligence (AI) based systems ability to comprehend and generate human-like text \cite{gao2023retrieval, lewis2020retrieval}. By combining a sophisticated retrieval mechanism with a powerful generation model, RAG systems can produce detailed, contextually relevant responses that significantly improve standalone language models limitations in terms of precision and human-like generation. The integration of semantic search into RAG systems is crucial, particularly for processing complex queries or those requiring deep contextual understanding, making it a cornerstone for enhancing retrieval accuracy and the quality of generated content.

Similar to the majority of research endeavors and NLP tasks, the Arabic language semantic search and RAG lags behind other languages due to the challenges posed by the Arabic language, including its complex morphology the diversity of its dialects and the shortage of datasets \cite{shaalan2019challenges, guellil2021arabic}. The above-mentioned challenges underscore the need for NLP techniques and studies tailored to Arabic in the context of semantic search and RAG systems. This paper aims to evaluate semantic search\'s effectiveness in processing Arabic alongside its impact on the performance of RAG systems specifically designed for Arabic Question Answering use-case. By evaluating the effect of different text encoders on the performance of RAG systems, the study seeks to provide insights into optimizing NLP applications for Arabic-speaking users and advancing the development of linguistically inclusive AI systems.

The rest of the paper is presented as follows; section 2 presents a simple overview of the previous literature, followed by section 3 which describes the methodology and experiment designed to evaluate the semantic search in context of RAG, while section 4 presents the results and the discussion finally followed by the conclusion.

\section{Literature review}

The evolution of semantic search can be traced back to before the widespread adoption of machine learning, initially relying on keyword-based methods and statistical techniques like latent semantic indexing (LSI) in the late 1990s \cite{hofmann1999probabilistic}. These early methods aimed to understand document similarity beyond exact keyword matches, setting the stage for more sophisticated approaches.

In the 2000s, machine learning transformed semantic search with algorithms like support vector machines (SVMs)\cite{lee2005classifier}, moving beyond keyword matching to a deeper understanding of queries through the use of more advanced and sophisticated textual features. However, these techniques had their limitations regarding the contextual meaning it is capable of grasping, therefore more advanced technique that are capable of leveraging the increasing influx of data were required.

Therefore, deep learning was introduced to semantic search which marked a significant milestone, with techniques like Word2Vec \cite{mikolov2013efficient} and GloVe \cite{pennington2014glove} enhancing the understanding of word relationships through unsupervised learning of huge corpus. This era also saw the development of advanced neural networks based architecture, such as attention mechanism \cite{vaswani2017attention} and transformers (e.g., BERT, GPT)\cite{devlin2018bert}, revolutionizing the field by enabling a deeper grasp of query intent and contextual relevance, which played a pivotal role the advancements of deep learning.

Parallel to the advancements of text encoders using deep learning, the use of approximate nearest neighbors (ANN) techniques became crucial for semantic search \cite{li2019approximate}. ANN algorithms, emerging in the early 2010s, facilitated efficient similarity search in high-dimensional spaces, essential for managing the vast data processed by deep learning models, which allowed users to replace slow unscalable techniques such as cosine similarity.
The integration of ANN with state-of-the-art language models has continuously improved of semantic search, enhancing its scalability and efficiency \cite{douze2024faiss}.

Significant progress has been made in Arabic semantic search thanks to deep learning technologies, especially with the creation of Arabic-focused encoders employing various frameworks like Word2Vec and Transformer-based models \cite{al2018deep, meddeb2021deep, al2015semantic}. Additionally, incorporating Arabic into universal encoders has also proved beneficial, utilizing patterns from other languages to improve understanding. This study highlights the effectiveness of using advanced deep learning methods to better grasp the semantic nuances of Arabic inquiries, leading to high precision in recognizing comparable questions in a customer support setting.

Conversely, RAG—Retrieved Augmented Generation—remains an emerging field within the burgeoning domain of generative artificial intelligence. The particular challenges associated with Arabic, such as its complex morphology and the relative paucity of resources, have hindered the attention and research it has received. Arabic RAG has not yet emerged as a focal point of scholarly inquiry to the degree that perhaps it warrants \cite{abdelazimsemantic, ram2023context}.

This study endeavors to assess the current landscape of semantic search capabilities within the Arabic language, addressing the notable lack of benchmarks and baseline data. Additionally, we seek to explore the impact of semantic search on the efficacy of retrieved augmented generation, positing that enhanced search mechanisms could significantly bolster the generation process.

\section{Evaluation Methodology}
In this section, we present the procedures employed for the evaluation of various semantic search modules. Our methodological framework is structured around three core components: dataset generation, evaluation metrics, and the configuration of semantic search modules.

\subsection{Dataset Generation}
The structure of the dataset required for the effective evaluation of semantic search ranking is pivotal. The dataset needs to encompass:

\begin{itemize}
    \item A collection of documents, which, in our study, consist of Arabic summaries of customer support calls for real-world companies.
    \item A set of user search queries, where each query is associated with all or a subset of the documents.
    \item A relevance score or label for each (query, document) pair, indicating the document\'s relevance to the query. This can be a binary label (relevant/irrelevant) or a score value with higher scores indicating higher relevance.
\end{itemize}

To circumvent the resource-intensive process of data collection and labeling, we leverage the capabilities of Large Language Models (LLMs), notably GPT-4, to generate search queries that:

\begin{itemize}
    \item Mimic realistic searches as might be conducted by customer support agents.
    \item Assign each query to a set of five summaries.
    \item Assign a relevance label/score to each (query, document) pair, with the scoring system of (irrelevant, somewhat relevant, very relevant) designated as (0, 1, 2) respectively.
\end{itemize}

A prompt was devised to meet these requirements and to generate an Arabic search query for every set of five summaries, ensuring that at least one summary is highly relevant.

The evaluation dataset comprises 2030 customer support call summaries and 406 search queries. A manual inspection of 10\% of the dataset (205 summaries with 41 queries) revealed only 2 misclassifications between very relevant and irrelevant summaries, suggesting the dataset is in a highly accurate state. Further validation on another set of randomized samples confirmed the dataset\'s robustness, qualifying it for semantic search evaluation.

\subsection{Evaluation Metrics}
Assessing the proficiency and effectiveness of various semantic search methodologies necessitates the use of specific evaluation metrics. Our attention centers on the pivotal metrics: Normalized Discounted Cumulative Gain (nDCG), Mean Reciprocal Rank (MRR), and Mean Average Precision (mAP). Each metric is integral to an in-depth evaluation of the ability of semantic search methods to accurately retrieve and rank documents. The formulas for calculating these metrics are delineated below:

\subsubsection{Normalized Discounted Cumulative Gain (nDCG)}

This metric quantifies the ranking efficacy across search outputs, considering the varying degrees of relevance each document holds. It takes into account the weighted relevance of documents, giving precedence to those at the top of the list. A superior nDCG value signifies that documents of relevance are appropriately prioritized in the rankings, thus underscoring the significance of their order in the search results. The nDCG is calculated using the following equation:

\begin{equation}
nDCG@k = \frac{DCG@k}{IDCG@k}
\end{equation}

where $DCG@k$ (Discounted Cumulative Gain at rank k) is defined as:

\begin{equation}
DCG@k = \sum_{i=1}^{k} \frac{2^{rel_i} - 1}{\log_2(i+1)}
\end{equation}

and $IDCG@k$ is the ideal DCG at k, representing the optimal ranking which maps to the maximum possible DCG up to position k, ensuring a fair comparison by normalizing the score. $rel_i$ represents the graded relevance of the result at position $i$.

\subsubsection{Mean Reciprocal Rank (MRR)}

MRR centers on the ranking of the first highly relevant document for a given search query, offering insights into the speed at which the ranking system is able to locate and display the most pertinent information. The calculation is as follows:

\begin{equation}
MRR = \frac{1}{|Q|} \sum_{i=1}^{|Q|} \frac{1}{rank_i}
\end{equation}

Here, $|Q|$ is the total number of queries, and $rank_i$ is the rank position of the first very relevant document for the $i^{th}$ query.

\subsubsection{Mean Average Precision (mAP)}

mAP serves as a comprehensive indicator of precision for all pertinent documents associated with each query, with an average taken across all queries. Higher mAP scores signify the system\'s enhanced consistency in identifying and retrieving relevant documents across the entire ranking.

The equation for the Average Precision (AP) for a single query is:
\begin{equation}
 AP = \frac{\sum_{k=1}^{n} (P(k) \times rel_k}{\text{\# relevant\_documents}}
\end{equation}

where:
- \( n \) is the number of retrieved documents.
- \( P(k) \) is the precision at cutoff \( k \) in the list of retrieved documents.
- \( rel_k \) is the relevancy score of the document at rank \( k \) whether it is very relevant, somewhat relevant or irrelevant.

To find the Mean Average Precision (MAP) we average the AP scores across all queries:

\begin{equation}
 MAP = \frac{\sum_{q=1}^{Q} AP_{q}}{Q} 
\end{equation}

where:
- \( AP_{q} \) is the Average Precision for the \( q^{th} \) query.
- \( Q \) is the total number of queries.

Employing these metrics allows for the evaluation of different methods' ranking effectiveness, taking into account essential aspects of accuracy and efficiency in document ranking that reflect their real-world use in semantic search scenarios.

\subsection{Semantic Search Approach}

As discussed in the Literature Review section, our evaluation framework is based on the idea of using encoders to convert documents and queries into embedding vectors that capture their content. We then determine the cosine similarity between the embedding vector of a query and those of the documents, ordering the documents based on these similarity scores.

\subsubsection{Assessment of Encoders}
The success of semantic search ranking relies significantly on the caliber of encoders used; higher-quality encoders produce more detailed embedding vectors, which in turn enable more accurate evaluations of similarity between search queries and documents. In our study, we chose encoders that show the best performance for Arabic by comparing their results against those obtained from random document rankings for each query (calculating average evaluation scores from 30 random rankings samples) and against the worst document rankings for each query (where documents are sorted by diminishing relevance to set a benchmark for the lowest achievable scores).

Evaluated encoders:
\begin{itemize} 

\item \textbf{Encoder \#1: Paraphrase Mulitlingual MiniLM\footnote{https://huggingface.co/sentence-transformers/paraphrase-multilingual-MiniLM-L12-v2}:}

This is a multi-lingual embedding model that was taught on 50+ languages covering Arabic and outputs a 384 dimensional embedding vector of the given sentence. It is mainly implemented for clustering and semantic search.

\item \textbf{Encoder \#2: Cmlm Multilingual \footnote{https://huggingface.co/sentence-transformers/use-cmlm-multilingual}:}

It is a universal sentence encoder, that is designed for mapping 109 languages to a common vector space. Leveraging LaBSE as its base model with embedding vector dimension of 768. Trained for multiple downstream tasks.

\item \textbf{Encoder \#3: Paraphrase Mulitlingual Mpnet \footnote{https://huggingface.co/sentence-transformers/paraphrase-multilingual-mpnet-base-v2}:}

This embedding model maps sentences \& paragraphs to a 768 dimensional dense vector space, and it works on 50+ languages including Arabic. Trained for tasks like clustering and semantic similarity.

\item \textbf{Encoder \#4: Multilingual Distil Bert \footnote{https://huggingface.co/sentence-transformers/distiluse-base-multilingual-cased-v1}:}

It is an embedding model that operates by mapping sentences to a dense vector space with a dimensionality of 512. It is effective across 15 languages, including Arabic. Can be used for tasks like clustering or semantic search.

\item \textbf{Encoder \#5: Xlm Roberta \footnote{https://huggingface.co/symanto/sn-xlm-roberta-base-snli-mnli-anli-xnli}:}

An embedding model that converts texts into a dense vector with 768 dimensions, it works well in Arabic and 12 other languages, it was trained on SNLI, MNLI, ANLI and XNLI corpora.
\end{itemize}

\subsection{RAG Evaluation Setup}

Retrieval-Augmented Generation for Arabic semantic search leverages both retrieval of relevant documents and generation of text to provide answers that are semantically aligned with the user\'s query. The complete RAG pipeline is shown in figure 1.

\begin{figure}[h
]
    \centering
    \includegraphics[width=\textwidth, height=7.5cm]{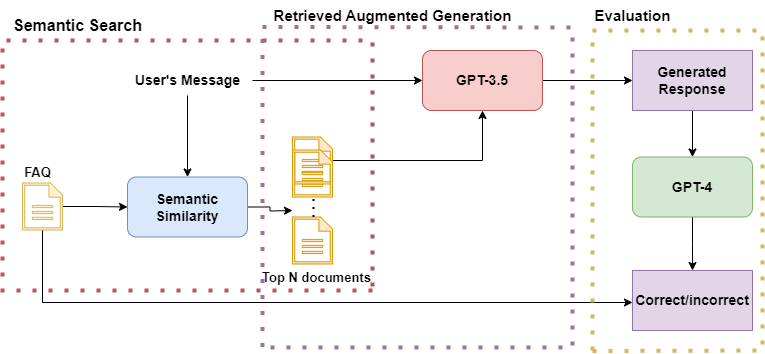}
    \caption{Retrieved-Augmented-Generation with Semantic Search Pipeline}
    \label{fig:enter-label}
\end{figure}

\subsubsection{Dataset Creation}

In constructing a comprehensive dataset for Retrieval Augmented Generation (RAG), we curated an extensive collection of Frequently Asked Questions (FAQs)from four different domains. This compilation contains a total of 816 distinct questions accompanied by their verifiable answers. To build a test set  and ensure its robustness, we employed the advanced capabilities of GPT-4, generating three nuanced variations for each original question. This approach yielded a synthesized dataset, optimally prepared for the subsequent testing of the Retrieval-Augmented Generation (RAG) framework.

\subsubsection{RAG Pipeline Implementation}

The RAG framework was precisely designed to undertake a multi-step process, aiming to test the semantic search capabilities within the Arabic language domain. This process is outlined as follows:

\begin{enumerate}
    \item \textbf{Semantic Encoding:} Each generated query undergoes a semantic encoding process alongside all ground truth questions of the same domain. This step utilizes a semantic search encoder to identify and retrieve the three closest semantically relative to the input query.
    
    \item \textbf{Knowledge-Based Answer Generation:} The answers corresponding to the semantically aligned questions, along with the generated query, are presented to a Large Language Model (LLM). For this purpose, GPT-3.5-turbo was selected to generate responses based on the knowledge extracted from the answers of the identified questions. This phase emphasizes the model\'s ability to synthesize and repurpose existing knowledge to address previously unseen queries.
    
    \item \textbf{Assessment Phase:} To fact checking of the generated responses, a subsequent LLM, specifically GPT-4-turbo, evaluates each response. This step involves comparing the generated answer, the original query, and the ground truth answer, thereby assessing whether the generated response properly addresses the query with the same information as the ground truth answer.
    
    \item \textbf{Accuracy Calculation:} The final stage in the RAG evaluation framework is the quantification of accuracy. This metric is derived by calculating the proportion of queries for which the responses generated by the LLM were correct, based on the criteria established in the assessment phase. 
\end{enumerate}

This systematic evaluation presents potential applicability of semantic search methods and comparing their performance to the Arabic RAG pipeline.

\section{Results}

This section presents the results for the stand-alone evaluation of semantic search, followed by assessment of the RAG with the different encoders.

\subsection{Semantic Search Evaluation results}
\begin{table}[h]
\caption{Semantic Search Evaluation Results}
\centering
\begin{tabular}{ccccc}
\hline
\textbf{Model}          & \textbf{NDCG@3} & \textbf{MRR@3} & \textbf{mAP @3} & \textbf{Emb. Size} \\ \hline
\textbf{Encoder \#1}    & 0.853           & 0.888          & 0.863           & 384                \\
\textbf{Encoder \#2}    & 0.789           & 0.798          & 0.793           & 768                \\
\textbf{Encoder \#3}    & 0.879           & 0.911          & 0.888           & 768                \\
\textbf{Encoder \#4}    & 0.868           & 0.89           & 0.876           & 512                \\
\textbf{Encoder \#5}    & 0.837           & 0.848          & 0.854           & 768                \\
\textbf{Random Ranking} & 0.669           & 0.623          & 0.703           & —                  \\
\textbf{Worst Ranking}  & 0.32            & 0.138          & 0.401           & —                  \\ \hline
\end{tabular}
\end{table}
From results represented in Table 1, it is shown that Encoder \#3 (paraphrase-multilingual-mpnet-base-v2) is performing best for Arabic semantic search, however it has the largest embedding size which allow it to carry more semantic information, yet require more computational \& memory costs.
\subsection{Correlation between Semantic search accuracy and RAG}

\begin{table}[h]
\caption{RAG Using GPT-3.5 Evaluation}
\centering
\begin{tabular}{ccc}
\hline
\textbf{Encoder}        & \multicolumn{1}{l}{\textbf{Top 3 Accuracy}} & \multicolumn{1}{l}{\textbf{Top 1 Accuracy}} \\ \hline
\textbf{Encoder \#1}    & 59.31\%                                     & 61.15\%                                     \\
\textbf{Encoder \#2}    & 62.01\%                                     & 63.23\%                                     \\
\textbf{Encoder \#3}    & 63.11\%                                     & 63.84\%                                     \\
\textbf{Encoder \#4}    & 62.5\%                                      & 63.24\%                                     \\
\textbf{Encoder \#5}    & 57.84\%

  & N/A                                        \\
\textbf{Random Ranking} & 6.62\%

  & N/A                                        \\ \hline
\end{tabular}
\end{table}

In Table 2, some findings were observed. Initially, a minor decline in accuracy from top-1 to top-3 was detected, attributed to GPT-3.5 merging of information from the three outcomes of semantic search and failing to identify the correct answer for a subset of questions.

Furthermore, while Encoder \#1 demonstrated commendable performance in the Semantic Search Evaluation, its effectiveness diminished in the context of Retrieval Augmented Generation (RAG) evaluation. Contrarily, Encoder \#2 exhibited an inverse pattern of performance. This discrepancy can be attributed to the nature of the semantic search evaluation, which closely resembles an \textbf{Asymmetric Semantic Search} scenario, where embeddings of extensive text summaries and brief queries, averaging four words, are compared. On the other hand, the RAG scenario aligns more with a \textbf{Symmetric Semantic Search}, where the queries and the reference questions are nearly equal in length, thereby examining distinct facets and constraints of the semantic search component. The size of the embedding vector also plays a crucial role, as larger vectors can encapsulate more information, potentially enhancing overall performance, particularly for Arabic. This language poses greater challenges in language modeling compared to English.

\section{Conclusion}

The analysis presented above clearly demonstrates the viability and significance of incorporating semantic search into the Retrieval Augmented Generation (RAG) system. This integration has notably enhanced the quality and precision of generated content. Furthermore, employing semantic search for the retrieval of documents relevant to a query offers several advantages, such as the use of shorter prompts in terms of tokens. This not only contributes to more precise outcomes but also results in cost-effective and quicker inference. However, further investigations are still required to conclude that superior encoders lead to superior RAG results.
\bibliographystyle{unsrt}  
\bibliography{references}  

\end{document}